# Non-invasive maturity assessment of iPSC-CMs based on optical maturity characteristics using interpretable AI

Short title

Non-invasive maturity assessment of iPSC-CMs using interpretable AI

Fabian Scheurer[1,2,‡], Alexander Hammer[1,‡], Mario Schubert[3,‡], Robert-Patrick Steiner[3], Oliver Gamm[3], Kaomei Guan[3], Frank Sonntag[2], Hagen Malberg[1], Martin Schmidt[1,*]

[1] Institute of Biomedical Engineering, TU Dresden, Fetscherstr. 29, 01307 Dresden, Germany

[2] Fraunhofer Institute for Material and Beam Technology IWS, Winterbergstr. 28, 01277, Dresden, Germany

[3] Institute of Pharmacology and Toxicology, TU Dresden, Fetscherstr. 74, 01307 Dresden, Germany

‡ Equally contributing authors

* Corresponding author

Adress for correspondence:

Dr. Martin Schmidt

Institute of Biomedical Engineering, TU Dresden

Fetscherstr. 29, 01307 Dresden, Germany

E-Mail: martin.schmidt@tu-dresden.de

Phone: +49 351 463 39942

Fax: +49 351 463 36026

Non-invasive maturity assessment of iPSC-CMs using interpretable AI

**Abstract** (max. 250 words):

Human induced pluripotent stem cell-derived cardiomyocytes (iPSC-CMs) are an important resource for the identification of new therapeutic targets and cardioprotective drugs. After differentiation iPSC-CMs show an immature, fetal-like phenotype. Cultivation of iPSC-CMs in lipid-supplemented maturation medium (MM) strongly enhances their structural, metabolic and functional phenotype. Nevertheless, assessing iPSC-CM maturation state remains challenging as most methods are time consuming and go in line with cell damage or loss of the sample.

To address this issue, we developed a non-invasive approach for automated classification of iPSC-CM maturity through interpretable artificial intelligence (AI)-based analysis of beat characteristics derived from video-based motion analysis.

In a prospective study, we evaluated 230 video recordings of early-state, immature iPSC-CMs on day 21 after differentiation (d21) and more mature iPSC-CMs cultured in MM (d42, MM). For each recording, 10 features were extracted using Maia motion analysis software and entered into a support vector machine (SVM). The hyperparameters of the SVM were optimized in a grid search on 80 % of the data using 5-fold cross-validation.

The optimized model achieved an accuracy of 99.5 ± 1.1 % on a hold-out test set. Shapley Additive Explanations (SHAP) identified displacement, relaxation-rise time and beating duration as the most relevant features for assessing maturity level.

Our results suggest the use of non-invasive, optical motion analysis combined with AI-based methods as a tool to assess iPSC-CMs maturity and could be applied before performing functional readouts or drug testing. This may potentially reduce the variability and improve the reproducibility of experimental studies.

**Highlights** (3-5 bullet points):

- Non-invasive evaluation of the maturation state of human iPSC-CMs by analyzing spontaneous beating characteristics using interpretable AI.
- Distinguish immature from more mature iPSC-CMs with high accuracy of 99.5 % using a simple support vector machine.
- Methods of explainable artificial intelligence (xAI) such as Shapley Additive Explanations (SHAP) enable the identification of the most relevant beating characteristics to distinguish immature from the more mature iPSC-CMs.
- Our results suggest that non-invasive, optical evaluation of iPSC-CM maturation state may reduce experimental variability and improve the reproducibility of studies.

## 1. Introduction

The development of human induced pluripotent stem cell-derived cardiomyocytes (iPSC-CM) has the potential to advance both basic research and clinical applications [1,2]. However, the immature phenotype of iPSC-CMs in comparison to adult CMs represents a major limitation for their utilization in drug screenings or the recapitulation of clinical disease phenotypes in vitro [3–6].

Different methods have been described to enhance the structural, metabolic and functional development of iPSC-CMs, including lipid-supplementation of cultivation media (MM), patterning of the culture surface, optimized extracellular matrix composition, cultivation in 3D tissue models with other cardiac cell types, and electrostimulation (reviewed in [7]). Furthermore, recent studies highlight the relevance of the maturation state of iPSC-CMs for their sensitivity towards pathophysiological stimuli, such as hypoxia-induced cell death [6,8], and their response to cardioactive drugs [9–11]. In particular, an enhanced maturation of iPSC-CMs was shown to improve their potential to detect pro-arrhythmic effects and predict the safety margin of cardioactive drugs [10].

Due to this profound impact on the experimental results, evaluation of the maturation state of iPSC-CMs is crucial for experimental studies, in order to improve the translational potential of the results, to reduce biological variability between different experiments and thus to facilitate data interpretation [12].

The maturation state of iPSC-CMs is evaluated based on the analysis of structural, electrophysiological, functional, or metabolic cell properties as well as protein- and gene expression [13]. However, most of these techniques are time consuming, expensive and lead to loss of the cells as they require harvesting, fixation, replating, or the application of fluorescent dyes with long-term toxicity [14]. In contrast, the spontaneous beating activity of iPSC-CMs represents a functional hallmark of iPSC-CMs, which can be flexibly assessed during long-term culture and quantified via video-based motion analysis [15–17]. Using this technique, a broad set of different features describing the contractile activity can be obtained, including beating rate, contraction and relaxation time (C time and R time), maximum contraction and relaxation velocity (Max C and Max R), and beating duration (0–1 time). Importantly, recently published studies revealed changes in these beating parameters during maturation of iPSC-CMs [17–19].

Consequently, we hypothesized that video-based motion analysis of the beating activity of iPSC-CMs may hold the potential to assess their maturation state in a non-invasive manner. Moreover, it remains so far unclear which beating parameter or combination of beating parameters is best suited to estimate the maturation state of iPSC-CMs.

Artificial intelligence (AI), including machine learning or deep learning, offers great potential for interpretation of data from cell-based studies and has been used for drug cardiotoxicity testing [20] or the discrimination of atrial and ventricular cells [17], but not for the assessment of iPSC-CM maturity. A previous study demonstrated the application of supervised machine learning on multidimensional data from iPSC-CMs and engineered cardiac tissues subjected to drug exposure, enabling automated determination of cardioactive drugs and prediction of their mechanism of action [21].

Due to the lack of gold standards or necessary data, the use of AI is often directed toward simpler machine learning models. Support vector machines (SVMs) were commonly applied as learning models with remarkable results [12,22–24]. These models offer high generalizability, can be applied to smaller data sets, require few computational resources and are suitable for explanation models regarding the feature importance compared to more complex deep learning models. The interpretability of processed features is a key advantage of SVMs.

The objective of this study was to investigate the potential of interpretable AI, utilizing SVMs and explanation models, in differentiating between mature and immature iPSC-CMs based on the analysis of their beating parameters (Figure 1). In addition, we sought to identify the most suitable parameters for evaluating the maturation state of iPSC-CMs.

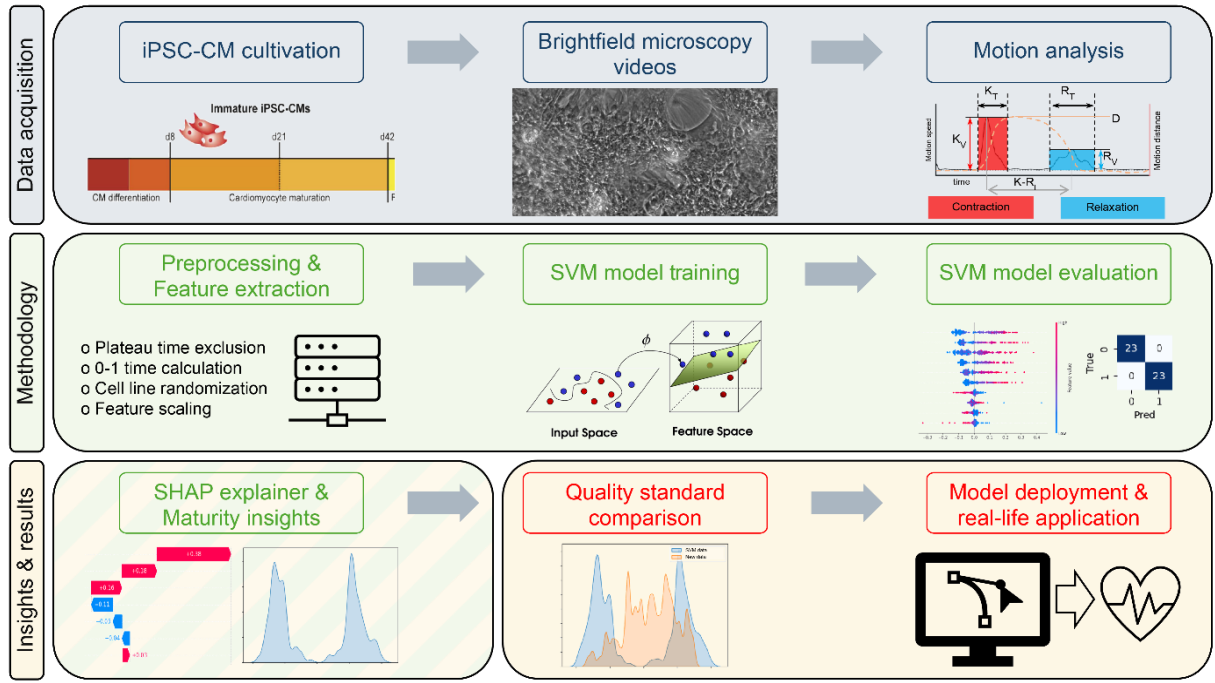

*Figure 1: Overview of iPSC-CM culture, data acquisition, model training for maturity classification, and analysis of individual parameter's influence on maturity classification. SHAP, Shapley additive explanations; SVM, support vector machine; iPSC-CM induced pluripotent stem cell-derived cardiomyocytes.*

## 2. Material and methods

### 2.1. *Data collection and iPSC-CM cultivation*

In this study, we used data of iPSC-lines generated from 4 healthy donors. iBM76 (UMGi005-A) and iWTD2 (UMGi001-A) cells were reprogrammed from mesenchymal stem cells and dermal fibroblasts, respectively, using STEMCCA lentivirus, and were characterized previously [25,26]. Cell lines isWT7 and isWT1 were reprogrammed from dermal fibroblasts using the integration-free CytoTune-iPS 2.0 Sendai Reprogramming Kit (Thermo Fisher Scientific) and characterized as previously described [11]. The generation of iPSCs was approved by the Ethics Committee of the University Medical Center Göttingen (approval number: 21/1/11 and 10/9/15) and performed according to the approval guidelines.

Directed cardiac differentiation was performed through modulation of Wnt signaling as previously described [27,28]. Briefly, iPSCs were grown to 85–85 % confluency on Geltrex-coated (Thermo Fisher Scientific) plates in Essential 8 medium (E8 medium, Thermo Fisher Scientific) with daily medium changes. Differentiation of iPSCs was performed with cardiac differentiation medium (C-diff), consisting of RPMI1640 with Glutamax and HEPES (Thermo Fisher Scientific), supplemented with 0.2 mg/ml ascorbic acid 2-phosphate (Sigma-Aldrich) and 0.5 mg/ml human recombinant albumin (Sigma-Aldrich). To initiate cardiac differentiation (day 0), E8 medium was replaced with C-diff supplemented with 4 µM CHIR99021 (Merck Millipore). After 48 hours, the medium was replaced with C-diff containing 5 µM IWP-2 (Merck Millipore) for an additional 48 hours. On day 6, the medium was changed to C-diff, and from day 8 onwards, cells were cultured in RPMI1640 with 2 % B27-supplement (B27 medium, Thermo Fisher Scientific). Medium was changed every other day. First contractions were observed at day 8–10. Between day 14–16, iPSC-CMs were replated into Geltrex-coated 6-well plates. Therefore, the medium was aspirated, and the cells were incubated in RPMI1640 with 1 mg/ml collagenase B (Worthington) for 30–60 minutes. Detached iPSC-CM sheets were transferred into 0.25 % trypsin/EDTA (Thermo Fisher Scientific) and incubated for 8 minutes at 37°C. Cells were resuspended in RPMI1640 with 20 % FBS and 2 µM thiazovivin (Merck Millipore), counted using a hemocytometer (Neubauer improved), and seeded into Geltrex-coated 6-well plates at a density of 0.5 million cells/well. After 24 hours, the medium was replaced with B27 medium. At day 21, iPSC-CMs were randomly assessed into 2 experimental groups cultured in B27 medium or MM, consisting of DMEM without glucose (Thermo Fisher Scientific) supplemented with 7 mM glucose (Sigma-Aldrich), 0.8 mM lactate (Sigma-Aldrich), 1.6 mM L-carnitine-hydrochloride (Sigma-Aldrich), 5 mM creatine-monohydrate (Sigma-Aldrich), 2 mM taurine (Sigma-Aldrich), 0.5 mM L-ascorbic acid-2-phosphate (Sigma-Aldrich), 0.5 % AlbuMax I lipid-rich BSA species (Thermo Fisher Scientific), 2% B27-supplement without insulin (Thermo Fisher Scientific), 50 nM human insulin (Sigma), 1 % Knockout Serum replacement (Gibco), 1 % non-essential amino acids (Gibco), 82 nM biotin (Sigma-Aldrich), and 0.37 nM vitamin B12 (Sigma-Aldrich). The culture medium was replaced every other day (2 mL per well).

*2.2. Video acquisition and motion analysis*

The spontaneous beating activity of iPSC-CM was recorded at different time points to assess iPSC-CMs at distinct maturation states (Figure 2 A). To determine the beating status of immature iPSC-CMs, video recordings were performed on day 21, before cultures were distributed into experimental groups receiving B27 medium or MM. Video data from iPSC-CMs at day 42, after 21 days in MM, were used to define the beating properties of cells with an enhanced maturation state, which was characterized previously based on a variety of different molecular and functional analyses [11].

For improved readability, we define iPSC-CMs, cultured in B27, to be immature on day 21 days we define designated as 'immature' and those cultured in MM are designated as 'mature' on day 42, recognizing that functional maturity may not be fully achieved under all conditions.

Two videos were recorded for each well/culture at different, randomly selected positions. Videos were obtained using an ORCA Flash 4.0 V3 CMOS camera (Hamamatsu Photonics) with a length of 20–30 seconds, a frame rate of 60 Hz and a resolution of 0.65 µm/pixel in an area of 670 × 670 µm (1024 × 1024 pixel). The videos were recorded in .czi raw format, transformed into MPEG4 format using ZEN black software (Carl Zeiss) and analyzed using the Maia software [28] using a block width of 10.4 µm, a frame offset of 67 ms and a maximum shift of 4.69 µm. Contraction and relaxation peaks as well as motion start and endpoints were manually assessed.

# Non-invasive maturity assessment of iPSC-CMs using interpretable AI

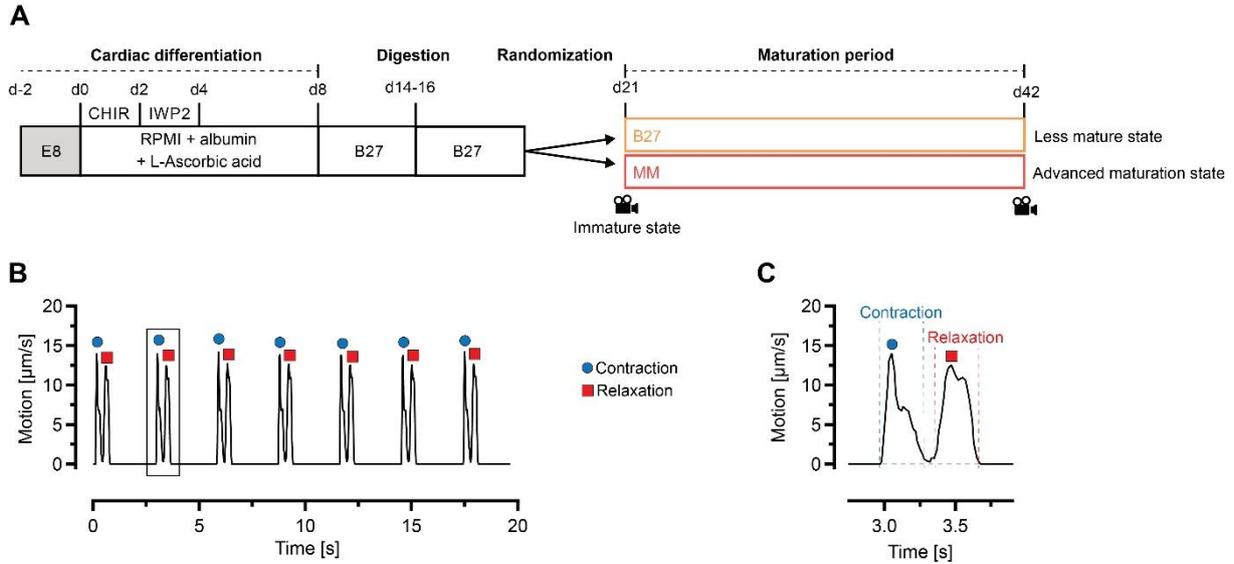

*Figure 2: Experimental design, data acquisition and analysis.* ***A****, Scheme of experimental groups and timepoints of video capturing (d21 and d42).* ***B****, Representative trace of iPSC-CM motion determined with Maia software.* ***C****, Magnification of region of interest marked in B showing one representative beating cycle, consisting of contraction and relaxation phase.*

## 2.3. Feature extraction

For each video, 10 features were extracted using the Maia motion analysis software as summarized in Figure 3. The feature time series, i.e. the representation of features over time, were z-normalized feature-wise to counteract strong differences in parameter expressions.

Due to the optimization of the Maia software during its development, in 33.1 % of all samples the 0–1 time was missing. Accordingly, the 0–1 time was calculated by adding together the contraction time, relaxation time and the mean plateau time for each cell line.

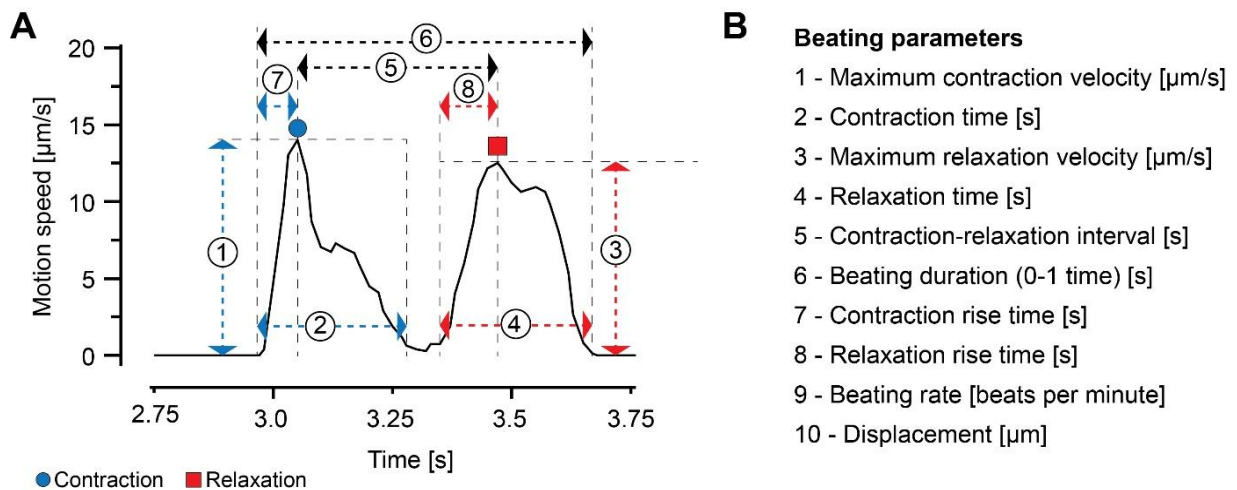

*Figure 3: Extraction of parameters/features from video data. Representative beating cycle (****A****) and list of extracted parameters (****B****). Displacement was calculated as the integral of the movement speed during the contraction phase. Maxima as well as start and end points of contraction and relaxation were manually assessed.*

Data preparation and all following operations were performed using Jupyter Notebook running on a Python 3 kernel. The scikit-learn package [29] version 1.2.2 was used.

## 2.4. Data splits

The overall dataset contained 362 videos of which 130 videos derived from immature iPSC-CMs (day 20/21, from 10 independent differentiations of 4 iPSC lines), 115 videos derived from iPSC-CMs with an improved maturation state cultured in MM (day 42, MM from 14 independent differentiations of 4 iPSC lines), and 117 were obtained from iPSC-CMs with a lower maturation state cultured in B27 medium (day 42, B27, from 16 independent differentiations of 4 iPSC lines).

Given the experimental setup and video capturing, mature cells (n = 115 videos, MM at day 42) were less represented than immature cells (n = 130 videos, day 21). Furthermore, the different cell lines were not equally represented at each maturity level. Therefore, we picked n = 230 recordings pseudo-randomly to equalize the number of recordings per maturity level (115 recordings per maturity level), considering an equal distribution of the represented cell lines. The data was partitioned into a training/validation set and a test data set in the ratio of 80 to 20 as summarized in Table 1. Thus, 184 samples (92 videos each from iPSC-CMs at day 21 and in MM at day 42) were employed for training and validation in a 5-fold cross-validation, which was repeated multiple times in a grid search for hyperparameter optimization. The final model was tested on the 46 retained samples. The partitioning was pseudo-randomized (random seed 42) and stratified, considering balanced classes and homogeneously distributed cell lines.

*Table 1: Data distribution per split regarding date of recording, day 21 (cultured in B27 medium) and day 42 (cultured in MM), and cell line.*

| Data split | Day 21, B27 | | Day 42, MM | | Total | |
|---|---|---|---|---|---|---|
| | *n* | *Cell lines* | *n* | *Cell lines* | *n* | *Cell lines* |
| **Training/validation set** *5-fold cross-validation* | 92 | 4 | 92 | 4 | 184 | 4 |
| **Test set** *Hold-out* | 23 | 4 | 23 | 4 | 46 | 4 |
| **Total** | 115 | | 115 | | 230 | |

## 2.5. Maturity classification model development

An SVM model separates two classes by calculating the offset and direction of a plane, in our application the mature and immature cell cultures being on opposite sites. It does so in maximizing a margin $m$, as part of the plane's direction vector $\vec{w}$ around this hyperplane in an $n$-dimensional space. By introducing a slack variable $\xi_i$, the model can be further optimized in a way to tolerate classification errors. The variable $C$ controls the way misclassifications are tolerated according to equation (1) [30]. The minimization term $0.5 \cdot \|\vec{w}\|^2$ is added with the product of the regularization variable $C$ times the sum of the slack variables $\xi_i$ for every sample [29–31].

$$\vec{w}, t, \xi_i = \arg \min_{\vec{w}, t, \xi_i} \frac{1}{2} \cdot \|\vec{w}\|^2 + C \cdot \sum_{i=1}^{n} \xi_i \qquad (1)$$

An instance of a SVM model was created. After fitting the model with the training set, it was analyzed using the confusion matrix and its derived evaluation functions from scikit-learn in Python.

For hyperparameter optimization, an initial random search was performed, followed by a detailed grid search around the hyperparameter combination resulting from random search. In the initial random search, the optimal precision through alteration of hyperparameters was searched for in 200 iterations. This included the regularization parameters *C* (reciprocal(0.1, 3)), different *kernel* functions (radial basis function, polynomial, sigmoid), the influence of a single training example *gamma* (auto, scale), the *degree* of the polynomial kernel (2, 3, 4), and the offset parameter *coef0* (uniform(0,1)). During this process, training and test parameters are recorded and evaluated. The parameters from the best model were then used as the starting point for the grid search.

In the subsequent grid search, the model was optimized by varying 5 hyperparameters, including *C* (0.1 : 0.1 : 1), *kernel* (radial basis function, polynomial), *gamma* (auto, scale), *degree* (2, 3), and *coef0* (0 : 0.05 : 1). We used accuracy as the optimization parameter in a 5-fold cross-validation.

*2.6. Model evaluation*

Model accuracy evaluation: The models from the 5-fold cross-validation with optimized hyperparameters were applied to the hold-out test set. Subsequently, another 5-fold cross-validation was performed on the full dataset (n = 230) with the same hyperparameter combination to account for the small dataset. In addition to the accuracy, we calculated the precision, the recall, and the F1-score as the mean across all folds, respectively. To estimate the robustness of the model, we additionally calculate the standard deviation (SD) of the metrics across all folds [32].

Evaluation of class separability: The underlying structure of SVMs as geometric models enables further analysis on the separability of classes in the feature space. The distance from the hyperplane that separates the two classes might indicate the degree of maturity of a cell culture. Therefore, we investigate the separability of both classes, using the decision function which is implemented in the scikit-learn package for Python [29].

Evaluation of feature relevance: We investigate the feature relevance for the automated classification decision and compare it with cell biological *a priori* knowledge. To explain the SVM model and obtain feature relevance values, we used Shapley Additive Explanation (SHAP) [33]. SHAP is a model agnostic approach that quantifies the contribution $\phi_{val}$ of each individual feature **F** to the classification decision. This provides an explanation of how the presence of the values $i$ of the features in any parameter selection and composition $S \subseteq \mathbf{F} \setminus \{i\}$, affect the model prediction. The average contribution of $i$ from the selection $S$ is normalized and summed up for all possible parameter combinations, as seen in equation (2) [33].

$$\phi_{val}(i) = \sum_{S \subseteq \mathbf{F} \setminus \{i\}} \frac{|S|!\,(p - |S| - 1)!}{p!} \Delta val(i, S) \qquad (2)$$

Non-invasive maturity assessment of iPSC-CMs using interpretable AI

## 3. Results

Beating parameters, extracted by using video-based motion analysis were used to train a SVM model to discriminate between immature and mature iPSC-CMs. Cells cultured in MM served as the matured state in our study based on previous studies demonstrating an enhanced maturation state of iPSC-CMs in MM in comparison to the widely used B27 medium with respect to electrophysiological function, calcium handling, metabolic activity and changes in gene expression [11]. Comparison of the beating features between both groups revealed strong differences between immature iPSC-CMs at day 20/21 and mature iPSC-CMs cultured in MM at day 42 (Figure 4). Mature cells showed significantly increased Max C, Max R and Displacement ($p < 0.0001$), prolonged C time, R time, C-R interval, beating duration, and R-rise time ($p < 0.0001$), shortened C-rise time ($p = 0.0005$) and a reduced spontaneous beating rate ($p = 0.0134$).

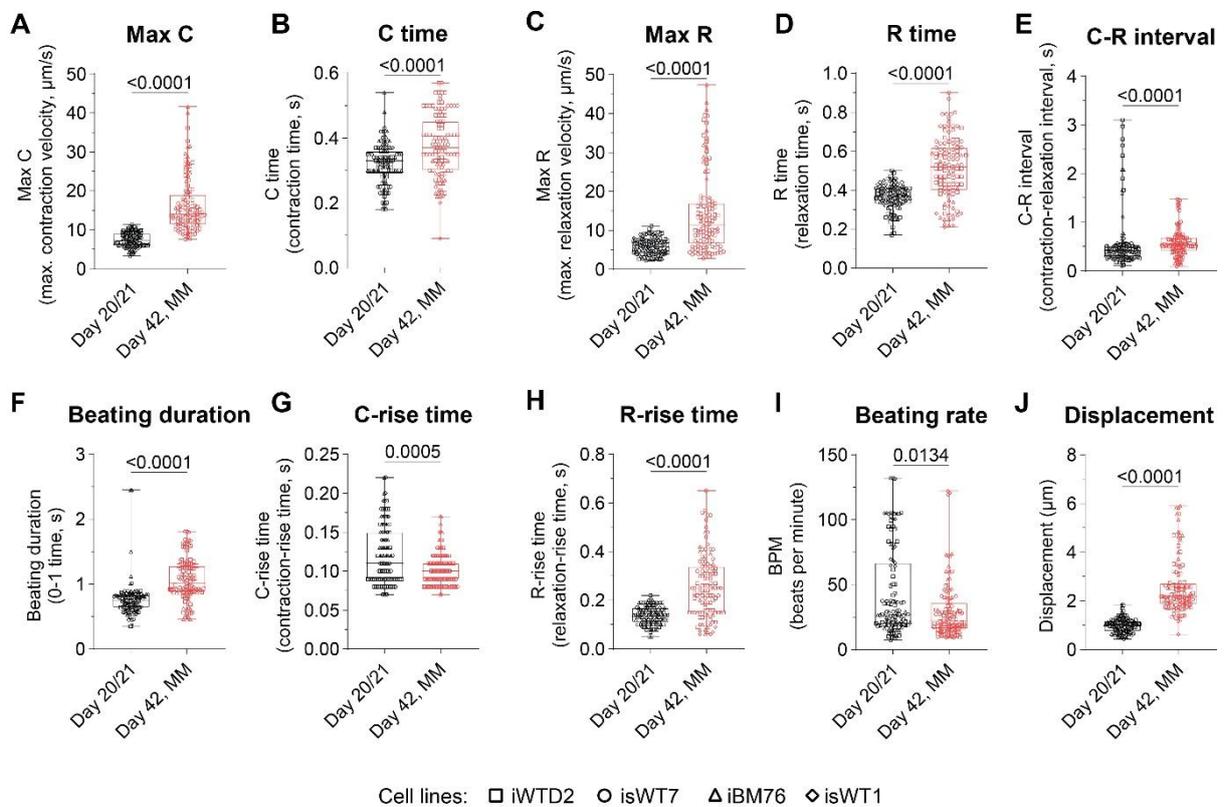

*Figure 4: Beating characteristics of early-state, immature iPSC-CMs at day 21 and mature iPSC-CMs after cultivation in MM at day 42. A-J; **A**, Maximum contraction velocity (Max C); **B**, contraction time (C time); **C**, maximum relaxation velocity (Max R); **D**, relaxation time (R time); **E**, contraction-relaxation interval (C-R interval); **F**, beating duration; **G**, contraction-rise time (C-rise time); **H**, relaxation-rise time (R-rise time); **I**, spontaneous beating rate; **J**, displacement. Data from n = 230 videos observed from 10 (day 21; $n_{21} = 115$) or 14 (day 42, $n_{42,MM} = 115$) independent differentiations of iPSC-CMs derived from iPSC-lines of 4 different donors (iWTD2, isWT7, iBM76, isWT1, indicated by symbols). Statistical analysis was performed using Kolmogorov-Smirnov test (unpaired, two-tailed).*

After hyperparameter optimization, the models scored on the test set with an accuracy of 99.5 ± 1.1 % (mean ± SD). The precision achieved on the test set was 100.0 ± 0.0 %, the recall was 98.8 ± 2.5 %, and the F1-score was 99.4 ± 1.3 %. The accuracy of the cross-validation on the full dataset was 99.1 ± 1.1 % the precision was 99.1 ± 1.9 %, the recall was 98.8 ± 2.5 %, and the F1-Score was 99.1 ± 1.1 % (Table 2).

# Non-invasive maturity assessment of iPSC-CMs using interpretable AI

*Table 2: Evaluation metrics results for test set, cross validation set and full data set on grid search model optimized for best accuracy. SD, standard deviation.*

|  | *n* | **Accuracy** mean ± SD | **Precision** mean ± SD | **Recall** mean ± SD | **F1-Score** mean ± SD |
|---|---|---|---|---|---|
| **Validation set** *5-fold cross-validation* | 184 | 99.3 ± 0.43 % | 100.0 ± 0.0 % | 98.6 ± 0.9 % | 99.3 ± 0.5 % |
| **Test set** *hold-out* | 46 | 99.5 ± 1.1 % | 100.0 ± 0.0 % | 98.8 ± 2.5 % | 99.4 ± 1.3 % |
| **Full data set** *5-fold cross-validation* | 230 | 99.1 ± 1.1 % | 99.1 ± 1.9 % | 99.2 ± 1.5 % | 99.1 ± 1.1 % |

The distribution of the SHAP values of the respective features used as inputs are displayed in Figure 5. The features were ranked by their impact on the classification outcome depending on the feature values. Samples in which the corresponding feature has a high impact on their classification outcome are assigned with high SHAP values. A positive SHAP value indicates that the corresponding feature is relevant for the detection of a class, whereas a negative SHAP value indicates that the corresponding feature is relevant for the rejection of this class. If the classification task is solved with high precision by a single feature and the model learns to solve the classification task using that feature, the absolute SHAP values are very large. A SHAP value of 0 indicates that the feature is not relevant for classification. Outliers might increase the range but do not affect the impact ranking. Each sample/video is color-coded according to the magnitude of the corresponding feature value, thus illustrating the degree to which a given feature is expressed in that instance. The effectiveness with which a sample can be classified using a feature is contingent upon the magnitude of the feature value. Consequently, a logical relationship is generally observed between the feature value and the SHAP value. In the context of a perfect relationship, a uniform color gradient is observed to form across samples of varying relevance.

The SHAP values of the feature displacement varied from –0.16 to 0.29, relaxation rise time from –0.13 to 0.31 and 0–1 time from –0.18 to 0.27 with an outlier at 0.41. Their ranges were 0.45, 0.44 and 0.45 respectively. The positive values were more stretched out, while the negative values were clustered closer together. The SHAP values of the Max C, the R time and the Max R showed a similar distribution. Max C ranging from –0.11 to 0.25, R time from –0.1 to 0.23 and Max R from –0.9 to 0.24. The SHAP value distribution of C-rise time, C-R interval and beating rate were clustered on the positive side, differing from the other features. C-rise-time ranges from –0.13 to 0.1, C-R-interval from –0.29 to 0.09 and beating rate from –0.09 to 0.07. The SHAP values of the C time were clustered around 0, ranging from –0.04 to 0.05.

Together, these results indicate that the first 6 features, displacement, R-rise time, 0–1 time, Max C, R time, and Max R, contributed to classify a mature culture, with higher feature values indicating a positive classification, and lower feature values indicating a negative classification (Figure 5). For these features, clustering occurred for samples with negative SHAP values, while samples with positive SHAP values were more widely scattered. For C-rise time, C-R interval, and beating rate, high feature values indicate against the classification of a mature culture, while lower feature values indicate for the classification. The clustering for these features is in positive SHAP values while negative SHAP values are more widely scattered. For C time, the SHAP values are very small and showed no relationship with the feature values, indicating that changes in C time do not express the maturity of iPSC-CMs and are therefore not relevant for the classification of immature and mature iPSC-CMs.

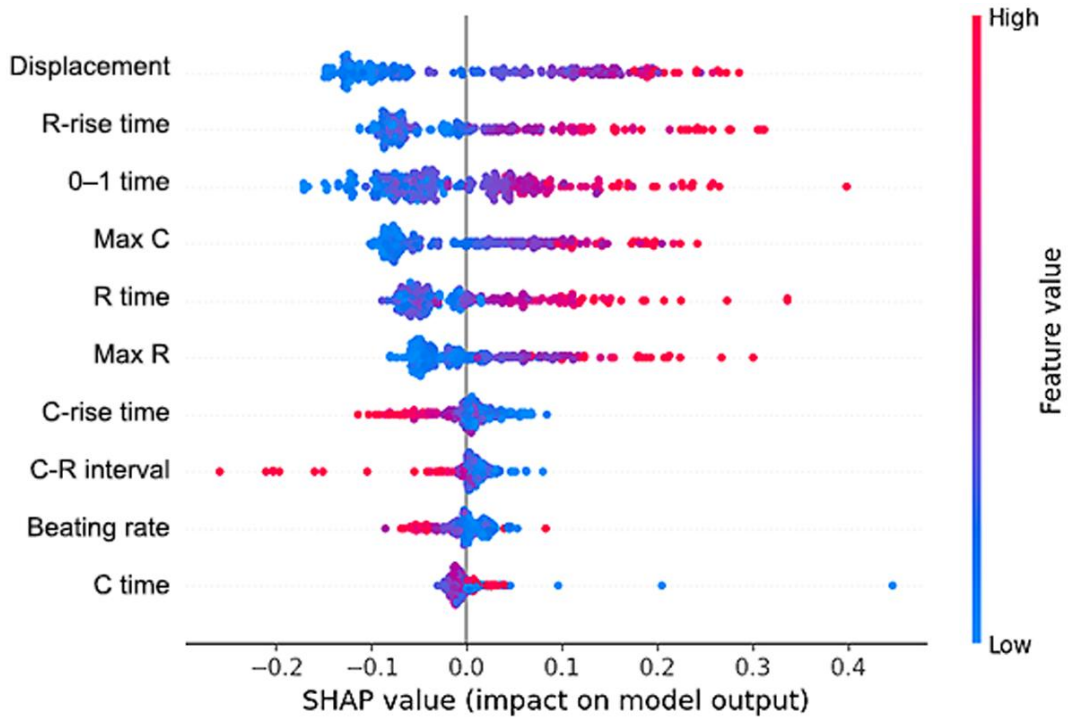

*Figure 5: SHAP beeswarm plot of the features that were used to train the model as well as their influence on the classification outcome. Each dot represents one sample (i.e. one video). The position of the samples on the abscissa indicates the relevance of each feature for the classification of each sample as immature or mature iPSC-CMs. High positive or negative SHAP values indicate that the feature is very well suited to class separation for the corresponding sample. The color-coding of the samples indicates the magnitude of corresponding feature value, from blue for low feature values to red for high feature values. In the case of a perfect relationship between feature values and SHAP values, the coloring of the samples with increasing or decreasing SHAP values of the corresponding feature corresponds exactly to the color gradient of the color bar.*

After successful establishment of the model, we included video data from iPSC-CMs that were cultured in B27 medium from day 21 until day 42. The iPSC-CMs cultured in B27 medium show a lower maturation state, as this medium mostly provides glucose as an energy substrate and only contains low levels of fatty acids [11]. Figure 6 displays SHAP waterfall plots for a random representative cell culture of A, immature iPSC-CMs at day 21, and B, mature iPSC-CMs cultured in MM medium at day 42 respectively, both classified correctly. The SHAP waterfall plots display explanations for individual predictions and thus allow insights into the individual feature importance for the classification of a sample as immature or mature iPSC-CMs. The 0–1 time and the features R-rise time and R time, correlating with the relaxation kinetics, showed the highest impact on the classification as matured state, whereas displacement and Max

Non-invasive maturity assessment of iPSC-CMs using interpretable AI

C had the strongest impact for classification into an immature state (Figure 6 A, B). Figure 6 C displays the SHAP waterfall plot for a random cell culture of iPSC-CMs cultured in B27 medium at day 42, classified as mature. In contrast to the examples presented in Figure 6 A and B, the features in this example are equally divided between the immaturity and the maturity of the iPSC-CM. The magnitude of the C-rise time appears to contradict the maturity of the iPSC-CM; however, the R-rise time, 0-1 time, and R time appear to support it.

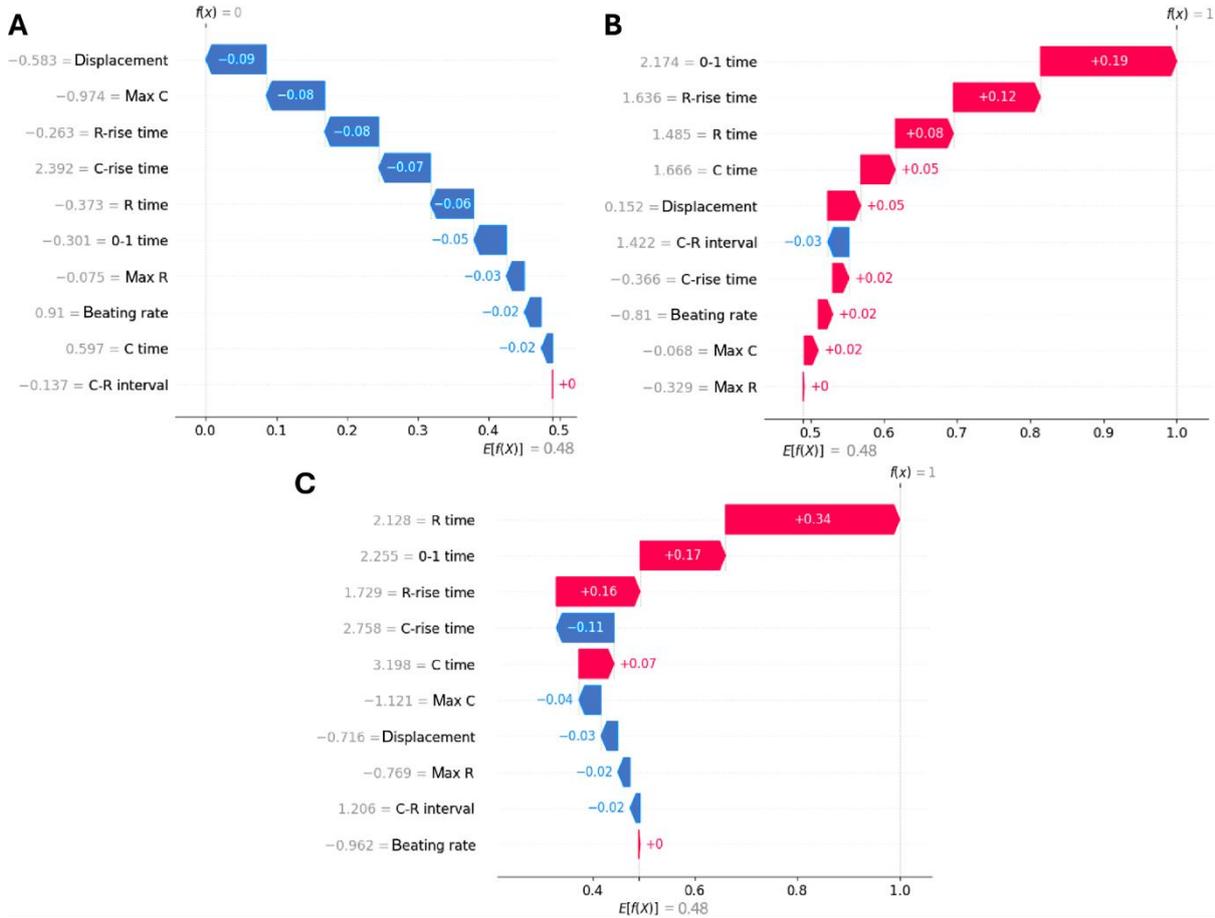

Figure 6: SHAP waterfall plots, showing the feature importance for individual predictions, for iPSC-CM cultures at different maturation states. **A**, Early state immature CMs at day 21. **B**, iPSC-CMs cultured in MM at day 42. **C**, iPSC-CMs cultured in B27 medium at day 42. The abscissa of the plots represents the expected classification outcome between immature ($f(x) = 0$) and mature ($f(x) = 1$). The contribution of the most relevant features to the classification result is shown line by line, from negative (blue) to positive (red) impact on the classification of an iPSC-CM as mature. The features are arranged in descending order of importance from top to bottom. The most relevant feature indicates the endpoint of the decision as either mature or immature. Each feature name is preceded by the corresponding normalized feature value.

Furthermore, the decision function calculated the distance of each sample from the model's hyperplane. The hyperplane is a separation plane that is placed in the feature space by the SVM to separate mature and immature cells. The 230 samples are clustered in two gauss-like distributions (Figure 7 A). The left cluster represents the immature cell cultures (day 20 iPSC-CMs), and the right cluster represents mature iPSC-CM cultures (day 42 in MM). The Kernel density plot shows a clear separation of the mature and immature samples, with no samples around the hyperplane at 0 (Figure 7 B). Samples with the same distance from the hyperplane tend to have similar feature expressions and therefore a comparable maturity.

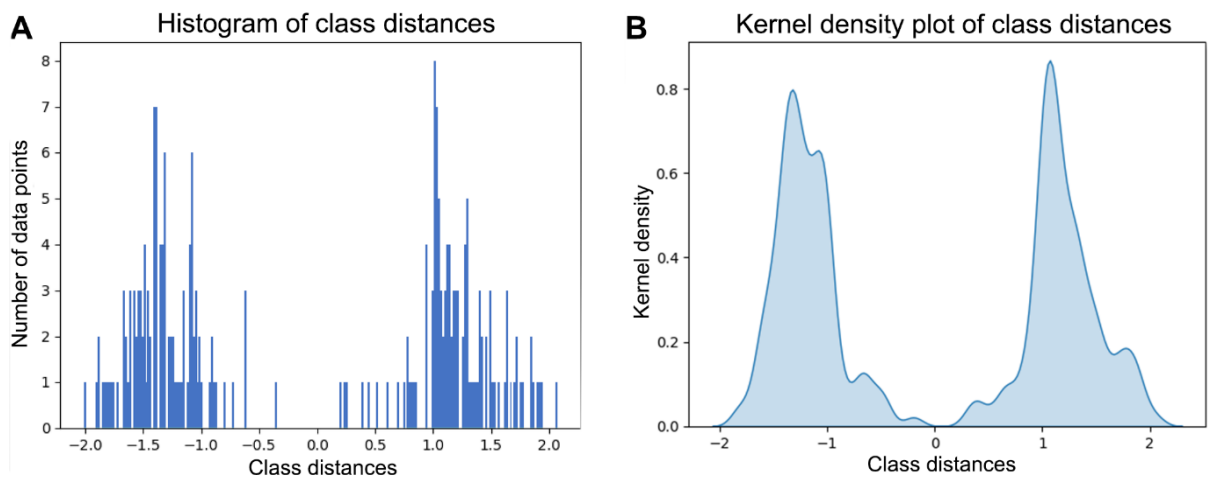

*Figure 7: Visualization of **A**, the histogram of class distances, showing the distribution of the distances of the samples from the hyperplane of the SVM and **B**, Kernel density plot, showing the smoothed probability density of the sample distances from the hyperplane SVM. The hyperplane is a separation plane that is placed in the feature space by the SVM to separate mature and immature cells. Samples in the positive value range are classified mature, samples in the negative value range are classified as immature.*

## 4. Discussion

Human iPSC-CMs represent invaluable models for preclinical research to characterize disease mechanisms, identify novel drug targets and as an experimental model to identify cardiotoxic or pro-arrhythmogenic substances [6,8]. However, the drug response of iPSC-CMs is strongly affected by their maturation state. Consequently, assessing the maturity of iPSC-CMs may be beneficial to enhance reproducibility of experimental studies and drug screenings. Video-based motion analysis provides a non-invasive method to characterize the contractile function based on various features related to the contraction as well as relaxation phase [15,28,34,35]. In this study, we aimed to investigate the potential of SVM models in assessing the maturation state of iPSC-CMs based on features/beating parameters obtained through video-based motion analysis. Beating data from iPSC-CMs with an improved maturation state through culture in lipid-enriched MM (day 42) and less mature iPSC-CMs early after differentiation (day 21) provided the basis for this analysis. Importantly, the enhanced maturation state of iPSC-CMs cultured in MM was proven based on a variety of functional and molecular analyses [11].

Our results show that even with this small data set, the SVM model is capable to discriminate between mature and immature iPSC-CM cell cultures with very good accuracy (above 99 %). Good generalization is achieved, and the underlying characteristics of the maturity pattern are recognized with respect to the extracted features. To ensure the robustness of the model, it is recommended to use cross-validation on the entire data set rather than relying solely on the hold-out test set. With a small test set population of 46 samples, the 5-fold cross-validation confirms the good accuracy and robustness of the model. The evaluation results showed a clear separation between the two classes, with only one or two misclassifications in different folds. These results indicate that machine learning methods can be applied to automatically evaluate the maturation state of iPSC-CM cultures using video-based beating features.

Investigation of the hyperplane of the SVM model, which represents the optimal data separation, revealed that some cultures that were classified as mature were close to the hyperplane. Further evaluation of the individual parameters of these iPSC-CM cultures indicated that this may be related to the iPSC-line and their donor-specific genetic backgrounds. It suggests that despite identical differentiation protocols and long-term maturation in MM, the feature values/expression varies depending on the iPSC-line donor. As the maturation state of the iPSC-CM cultures can be assessed flexibly during cultivation, the determination of the cell state using SVMs may provide information whether a similar maturation state is reached in different independent experiments, which potentially enhances reproducibility due to better synchronization of the cellular conditions.

When the SVM-model was applied to classify iPSC-CMs cultured in RPMI/B27 medium, we found that iPSC-CMs cultured in RPMI/B27 medium showed a heterogeneous distribution between the immature cluster on day 21 and the mature cluster in MM at day 42, indicating some maturation but a lower degree of maturity in comparison to iPSC-CMs cultured in MM (Figure 8). This is in line with our previous results demonstrating an enhanced maturation state of iPSC-CMs cultured in MM compared to B27 medium. Cells in MM showed an increased cell size, higher mitochondrial density, enhanced respiratory/metabolic activity, and enhanced calcium handling as well as electrophysiological properties [11]. Therefore, these results suggest that several levels of maturity can be distinguished using the SVM.

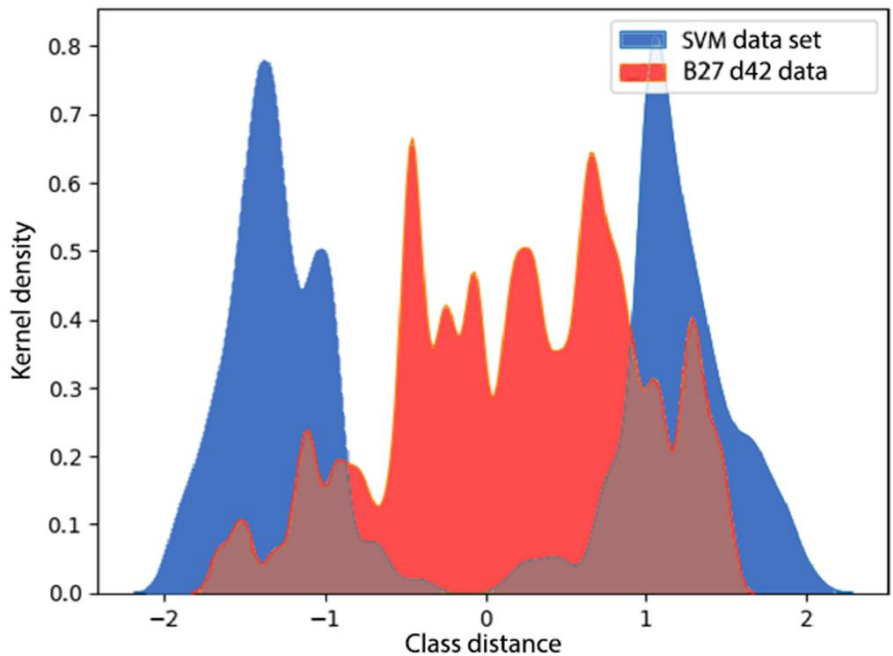

*Figure 8: Kernel density function for immature cell cultures on day 21 (left blue distribution), for mature cell lines cultured in MM medium on day 42 (right blue distribution), and for less mature cell lines cultured in B27 medium (red distribution). Blue colored distributions reflect the data set used for the training and validation process of the support vector machine (SVM).*

Next, we aimed to examine which feature/beating parameter or combination of features represent indicators to evaluate the maturation state of iPSC-CMs. The SHAP explanations revealed positive values for Displacement, R-rise time, 0–1 time, Max C and R time, which demonstrates that those are important for the classification of matured cell cultures (Figure 5). In contrast, C-rise time, C-R interval, beating rate, and contraction time tend to have negative SHAP values. The positive SHAP values of these features were concentrated in a narrow range, with a mean value below 0.1, and the contraction time exhibits outliers, indicating that these features had a limited impact on the model's decision-making process to classify immature and mature iPSC-CMs. Therefore, we identified that Displacement, R-rise time, 0–1 time, Max C and R time were most relevant for maturity classification of iPSC-CMs. However, future studies are required to clarify how these beating features correlate with the molecular cellular properties associated with iPSC-CM maturation, especially with respect to calcium handling and electrophysiology.

The improved calcium handling of iPSC-CMs during maturation in fatty-acid supplemented media is reflected by reduced diastolic calcium levels [4,11], increased transient amplitude [11,36,37] faster upstroke and decay kinetics [4,11,36,37]. We hypothesize that the enhanced calcium upstroke and decay kinetics observed in iPSC-CMs cultured in MM [11] may be related to R-time and beating parameters C-rise time, C-time and R-rise time, respectively [34], see Supplementary Figure 1. In addition, the calcium transient duration may correlate with the 0–1 tim, as indicated by previous studies which reported shortened calcium transient durations along with shortened contraction duration (measured based on impedance) [38,39]. Furthermore, linear correlations were reported between Max C (maximum contraction velocity) and contraction amplitude [35], and between displacement and traction force [39].

In addition to changes in calcium handling, iPSC-CMs cultured in MM showed remarkable differences with respect to action potential properties and ion currents. Hayakawa et al. simultaneously measured field potentials and motion of iPSC-CMs and found strong correlations between their field potential duration (FPD) and contraction duration under non-arrhythmic conditions [39]. In comparison to iPSC-CMs in B27 medium, iPSC-CMs cultured in MM showed increased densities of the sodium current ($I_{Na}$), the transient outward potassium current ($I_{to}$) and the inward rectifying potassium current ($I_{K1}$). In contrast, the L-type calcium current ($I_{CaL}$) was reduced in iPSC-CMs in MM compared to B27 medium [11]. This is in line with previous studies that found altered ion channel expression levels and current densities in iPSC-CMs compared to adult CMs. Adult CMs showed increased expression level/density of the sodium channel SCN5A/$I_{Na}$, the hERG channel KCNH2/$I_{Kr}$, and the potassium channel KCNQ/$I_{K1}$, while the expression/density of the L-type calcium channel CANA1C/$I_{CaL}$ were reduced compared to iPSC-CMs [40]. These electrophysiological differences between adult CMs and immature iPSC-CMs may contribute to the discrepancies that were observed for the prediction of cardiotoxic or pre-arrhythmic activity using iPSC-CMs in comparison to the actual risk profile observed in the clinic [10,40].

Limitations:

In our study, iPSC-CMs cultured in MM were assumed to be mature on day 42, but the maturation state of iPSC-CMs is still very different compared to adult CMs. However, fatty acid-supplementation of the culture medium represents a simple and cost-efficient approach without the need for specific materials, bioprinting or tissue engineering, and was already shown to improve electrophysiological development and drug response of iPSC-CMs. Similarly, video-based motion analysis is a simple, non-invasive and cost-effective method to assess the contractile activity of iPSC-CMs. Therefore, we believe that SVM models like ours can be widely applied in labs working with iPSC-CMs. The video data used to train the SVM were obtained from iPSC-CMs derived from iPSCs lines of 4 healthy donors. Although a robust classification was observed for all cell lines, it is unclear whether the SVM is capable of correctly classifying iPSC-CMs from other donors with distinct genetic background. Furthermore, future studies will focus on extracting the characteristics directly from the video data in order to account for further features that have not yet been considered in our current model, such as the direction of motion or degree of alignment of the motion vectors.

## 5. Conclusion

In this paper, we implemented a SVM that effectively distinguishes between immature early state (day 21) and mature (day 42) iPSC-CM cultures using beating characteristics. The model achieved an accuracy of 99.5 % on a hold-out test set, demonstrating highly accurate classification [41]. The SHAP explanations reveal displacement, R-rise time, 0–1 time, and Max C to be the most impactful beating characteristics for discriminating immature and mature iPSC-CMs. The application of the SVM model to classify iPSC-CMs cultured in B27 medium further revealed the lower maturation state of those compared to iPSC-CMs in MM, consistent with the findings of our previous study in which compared both groups using a variety experimental techniques [11]. Therefore, this non-invasive approach allows the evaluation of the maturation state of iPSC-CM cultures before testing of drugs or complex molecular/functional readouts, which may enhance the reproducibility between independent experiments and variation between iPSC-CMs derived from iPSCs of different donors.

## Author contributions

FS: Methodology, Software, Validation, Formal analysis, Investigation, Writing - Original Draft, Writing - Review & Editing, Visualization

AH: Methodology, Software, Validation, Formal analysis, Investigation, Writing - Original Draft, Writing - Review & Editing, Visualization, Supervision, Project administration

MSchu: Conceptualization, Methodology, Formal analysis, Investigation, Writing - Original Draft, Writing - Review & Editing, Visualization, Supervision, Project administration

OG: Formal analysis, Investigation, Data Curation

RPS: Formal analysis, Investigation, Data Curation

KG: Conceptualization, Methodology, Investigation, Resources, Writing - Review & Editing, Supervision, Project administration

FSo: Conceptualization, Methodology, Investigation, Writing - Review & Editing, Supervision, Project administration

HM: Conceptualization, Methodology, Investigation, Writing - Review & Editing, Supervision, Project administration

MSchm: Conceptualization, Methodology, Validation, Formal analysis, Investigation, Writing - Original Draft, Writing - Review & Editing, Visualization, Supervision, Project administration

## Declaration of competing interest

We declare no Conflict of Interest.

## Acknowledgements

This study was partly supported by grants from the European Union's Horizon 2020 research and innovation program (TIMELY, No. 101017424 to A.H.), the European Union and co-financed from tax revenues on the basis of the budget adopted by the Sexon State Parliament (Dynamic Resources, No. 100694616 to A.H.), the German Federal Institute for Risk Assessment (No. 60-0102-01#00067 - P639 to M.Schu.), the Free State of Saxony and the European Union (SAB EFRE projects HERMES with project number 100328421 to K.G. and F.So. as well as ESF Plus project "MultiMOD" with project number 100649621 to K.G.), the Deutsche Forschungsgemeinschaft (DFG, German Research Foundation) under project number 288034826 –IRTG 2251: "Immunological and Cellular Strategies in Metabolic Disease" to K.G, and the Medical Faculty of the TU Dresden (MeDDrive project) to M.Schu. O.G. is supported by the ESF-Plus PhD Scholarship Program. This project is co-funded by the European Union and co-financed from tax revenues on the basis of the budget adopted by the Saxon State Parliament. The Article Processing Charge (APC) were funded by the joint publication funds of the TU Dresden, including Carl Gustav Carus Faculty of Medicine, and the SLUB Dresden as well as the Open Access Publication Funding of the DFG.